\PassOptionsToPackage{table}{xcolor}
\documentclass{article}
\usepackage{spconf,amsmath,amssymb,graphicx,hyperref}
\usepackage{booktabs}
\usepackage[most]{tcolorbox}

\usepackage{microtype}

\usepackage{inconsolata}
\usepackage{longtable}

\usepackage{graphicx}
\usepackage{subcaption}
\usepackage{amsmath}
\usepackage{amssymb}
\usepackage{pifont}
\usepackage{booktabs} 
\usepackage[linesnumbered,ruled,vlined]{algorithm2e}
\usepackage{multirow}
\usepackage{float}

\usepackage{times}
\usepackage{latexsym}
\usepackage{soul} 
\usepackage{xcolor} 

\definecolor{violet}{HTML}{BEBADA}
\definecolor{StudentColor}{HTML}{CCEBC5}

\newcommand{\myparagraph}[1]{\noindent\textbf{#1}}

\title{Speech Language Models for Under-Represented Languages: Insights from Wolof}

\name{Yaya Sy$^{\star \dagger}$ \qquad Dioula Doucouré$^{\star}$ \qquad Christophe Cerisara$^{\star}$ \qquad Irina illina$^{\star}$}

\address{$^{\star}$ LORIA, CNRS, Nancy, France \\
      $^{\dagger}$ Soynade Research}
\begin{document}
%
\maketitle
\begin{abstract}
We present our journey in training a speech language model for Wolof, an underrepresented language spoken in West Africa, and share key insights. We first emphasize the importance of collecting large-scale, spontaneous, high-quality unsupervised speech data, and show that continued pretraining HuBERT on this dataset outperforms both the base model and African-centric models on ASR. We then integrate this speech encoder into a Wolof LLM to train the first Speech LLM for this language, extending its capabilities to tasks such as speech translation. Furthermore, we explore training the Speech LLM to perform multi-step Chain-of-Thought before transcribing or translating. Our results show that the Speech LLM not only improves speech recognition but also performs well in speech translation. The models and the code will be openly shared.
\end{abstract}

\begin{figure*}[ht]
    \begin{minipage}[c]{0.48\linewidth}
        \centering
        \includegraphics[width=\textwidth]{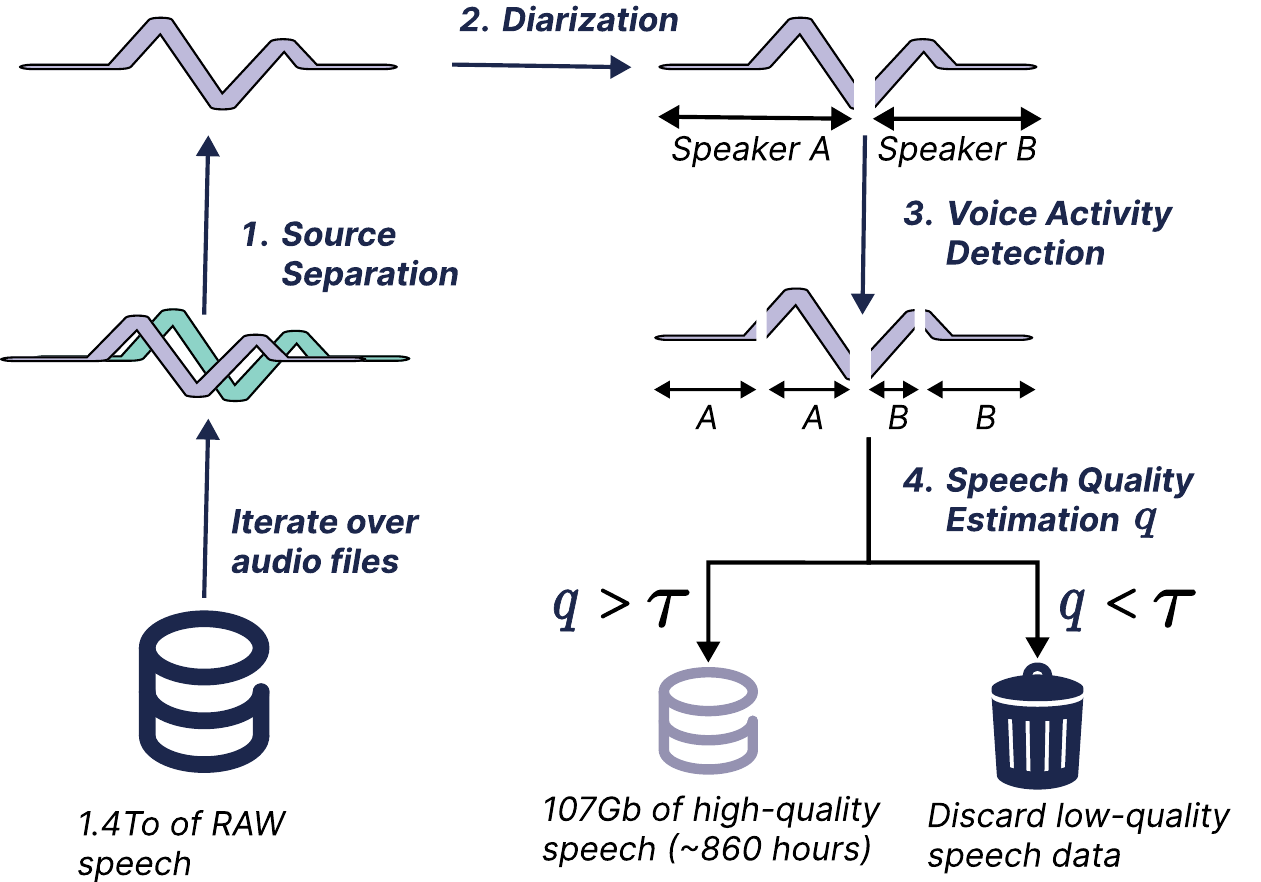}
        \caption{High-quality spontaneous speech dataset collection for self-supervised speech continued pretraining. We collected large scale raw Wolof speech on the internet and further refined it using multiple stages of processing: \textit{(1) Source Separation}, \textit{(2) Diarization}, (3) \textit{VAD}, and finally \textit{(4) Speech Quality Filtering} to filter low-quality speech utterances and only keep high-quality speech that will be used for continued pretraining.}
        \label{fig:data_pipeline}
    \end{minipage}
    \hfill
    \begin{minipage}[c]{0.48\linewidth}
        \begin{center}
            \includegraphics[width=\textwidth]{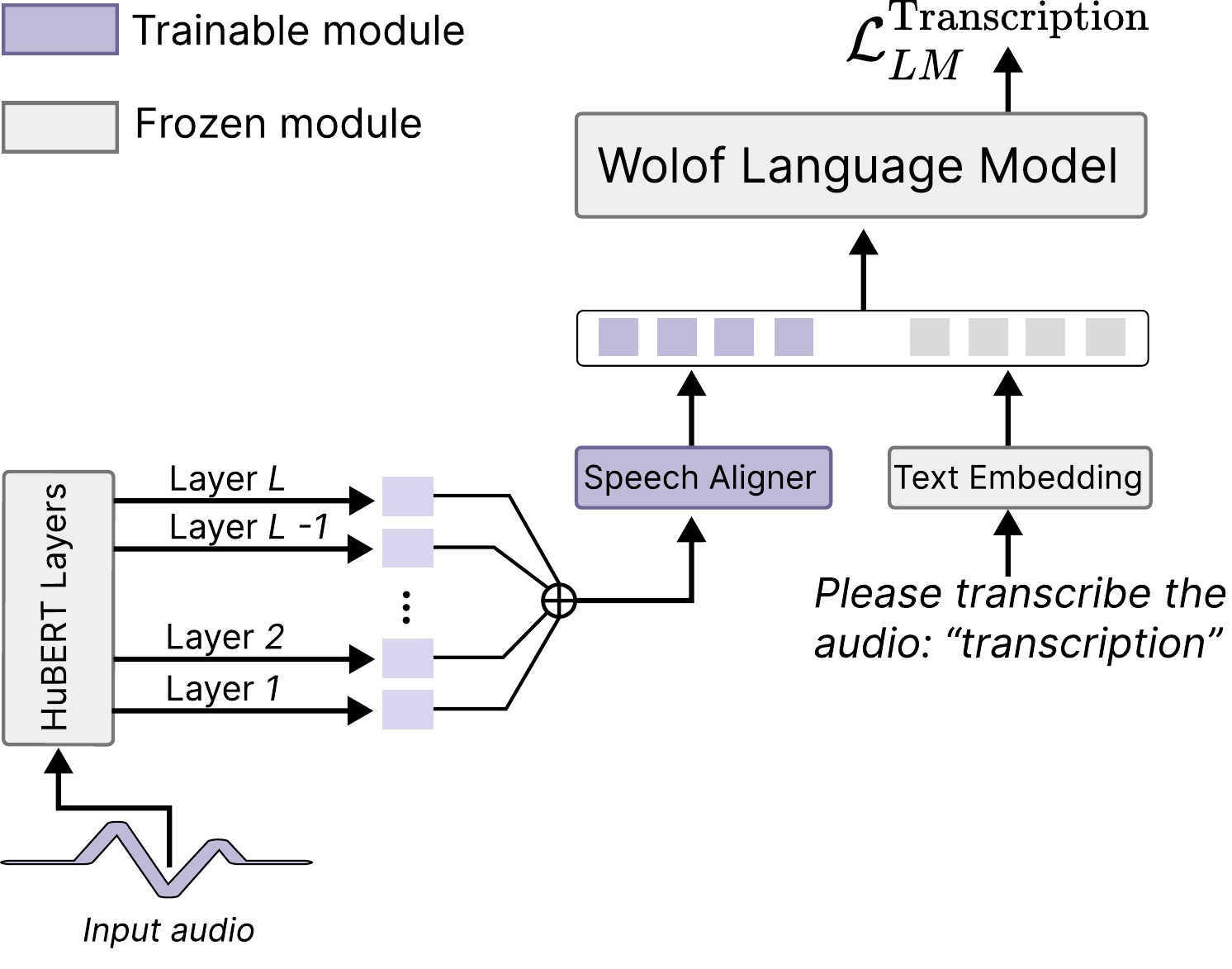}
            \captionsetup{width=\textwidth}
            \caption{The approach for integrating a speech encoder into an LLM. The speech features are extracted at different layers of the HuBERT model, concatenated ($\oplus$), projected by the speech aligner, and then finally fed to the LLM along with the text embedding. All modules are frozen during training, except the speech aligner.}
            \label{fig:speech_lm_approach}
        \end{center}
    \end{minipage}
\end{figure*}

\section{Introduction}
\label{sec:intro}

Self-supervised learning methods have become central to advances in speech models training, particularly for Low and Medium-Resource Languages, with models such as wav2vec \cite{wav2vec2} and HuBERT \cite{hubert} that learn speech representations from unlabeled audio.  Multilingual self-supervised learning has extended coverage accross languages, including recent efforts on African languages such as in \cite{gauthierSSL}, where the authors pretrained from scratch HuBERT on nearly 65k hours of speech. While such large-scale multilingual pretraining is a significant step, the performance on any given language is tied to its specific distribution within the pretraining data. Wolof, representing 0.1\% of the multilingual corpora in \cite{gauthierSSL}, there is room for substantial gains via language-specific training.

We address these limitations with three key contributions. First, we highlight the importance of training on spontaneous, language-specific data. To this end, we curated 1.4 TB of raw Wolof speech and filtered it to 860 hours of high-quality audio. Second, we show the effectiveness of continued pretraining, which allows to reuse the compute already invested in the base model. This strategy enables us to outperform both the original HuBERT checkpoint and the African-centric model on the ASR task, while requiring only 33 epochs compared to the hundreds of epochs needed to train the base HuBERT model. Third, we extend beyond speech recognition by integrating the speech model into a Wolof LLM. This not only improves ASR accuracy but also unlocks downstream capabilities such as speech-to-text translation

\section{Related Works}
In \myparagraph{Self-supervised Speech Representation Learning} frameworks such as Wav2Vec and HuBERT, the model first learns general-purpose speech representations during \textit{pretraining}, then fine-tuned for downstream tasks. A pioneering example is \textbf{Wav2Vec} \cite{wav2vec, wav2vec2}, which uses a contrastive loss to predict target tokens derived from discretized features produced by a convolutional encoder. However, the convolutional encoder may provide weak targets, and the contrastive loss requires careful design of negative samples. \textbf{HuBERT} \cite{hubert} addresses these issues by replacing the contrastive loss with a simpler cross-entropy objective and by generating the targets differently: instead of relying on the convolutional encoder, it applies KMeans clustering to the hidden representations from an intermediate transformer layer at a given training step. These clusters serve as discrete speech targets for the next training steps. \textbf{Data2Vec} \cite{baevski2023efficientselfsupervisedlearningcontextualized} further removes the discretization step by training the model to predict masked continuous features directly. In this work, we chose the HuBERT framework primarily for its simplicity and wide adoption.
\\
\\
\myparagraph{Multilingual Speech Models.} Many studies \cite{xlsr, mhubert} have expanded Wav2Vec or HuBERT to hundreds of languages, where the cross-lingual transfer benefits low-resource languages. Other efforts specifically target African languages \cite{gauthierSSL, afrihubert}. While these approaches are an important step for low-resource speech recognition, the downstream ASR performance for any given language depends on its representation in the pretraining data. Since self-supervised learning does not require labeled data, we advocate for large-scale, language-specific data collection.
\\
\\
\myparagraph{Speech Language Models.} Integrating a speech encoder into an LLM \cite{slm_methods, llamaspeech} enables cross-task generalization of the knowledge encoded in the language model. Since an LLM can be trained on monolingual, unlabeled text, this approach is particularly beneficial in low-resource scenarios. We show that integrating the Wolof HuBERT into a Wolof LLM we trained not only improves ASR performance but also unlocks new capabilities in speech-to-text translation.

\section{self-supervised speech representation learning with continued pretraining}
\label{sec:hubert}

\subsection{Data}
\myparagraph{Large-scale 
data collection.} We manually curated 1.4TB of raw Wolof speech from public web sources, prioritizing natural, spontaneous speech from native speakers over read speech, which can contain artifacts such as hyper-articulation or artificial emotions. As illustrated in Figure \ref{fig:data_pipeline}, this raw audio was processed through a filtering pipeline inspired by Emilia-Pipeline \cite{emilia, emilialarge, amphion} that includes \textbf{\textit{Source Separation}}, \textbf{\textit{Speaker Diarization}} and \textbf{\textit{Voice Activity Detection}}. We retained only utterances between 3-30 seconds that have a DNSMOS \cite{dnsmos} quality score above 3.2, resulting in a final dataset of 860 hours of high-quality Wolof speech. 

\subsection{HuBERT continued pretraining}

\myparagraph{Model Initialization.} We continued pretraining of HuBERT-base \cite{hubert}, a 90m parameter model pretrained on 960 hours of speech from the LibriSpeech dataset \cite{librispeech}. We chose this model not only for its wide adoption but also for its open license, which allows us to redistribute our model freely—unlike the other African-centric HuBERT variants cited. We initialize the weights of the model from the already pretrained HuBERT-base, and start with a fresh optimizer state, as the original one was not released. Nevertheless, the fact that our training dataset is nearly as large as Hubert's pretraining dataset justifies the term "continued pretraining".
\\
\\
\myparagraph{Training.} We continued pretraining the HuBERT-base model on 32 A100 GPUs, using the same hyperparameters: 33 epochs with a learning rate of $0.0005$ and a batch size of $87.5$ seconds. Note that the model is trained for far fewer epochs than HuBERT-base, which was trained for a total of 528 epochs. This is explained by the continued pretraining: we benefit from previous training steps and don't need to train as long as the base checkpoint.
\begin{table}[ht]
    \centering
    {\small{
    \begin{tabular}{lcccccc}
        \toprule
        \textbf{\textit{Split}} & \textbf{Fleurs} & \textbf{Alfa} & \textbf{CV} & \textbf{Kallama} & \textbf{UB} & \textbf{Total} \\
        \midrule
        \textbf{\textit{Train}} & 8.72 & 16.13 & 34.97 & 33.60 & 4.52 & \textbf{97.94} \\
        \textbf{\textit{Test}} & 1.75 & 2.84 & 6.21 & 5.91 & 1.12 & \textbf{17.83} \\
        \bottomrule
    \end{tabular}
    }}
    \caption{The Wolof datasets used for ASR fine-tuning and their sizes in hours of speech.}
    \label{tab:asr_data}
\end{table}

\subsection{Evaluation on an ASR task}
\label{sec:asr_exps}
After continued pretraining, we finetune the model for speech recognition on a supervised dataset.
\\
\\
\myparagraph{Data.} For supervised Wolof ASR, we used (1) \textbf{Fleurs}, (2) \textbf{Alfa} \cite{gauthierSSL}, (3) common-voice (\textbf{CV}), (4) \textbf{Kallama} \cite{kallama}, and (5) urban-bus \cite{urbanbus} (\textbf{UB}), a speech dataset of common places in Dakar city, the capital of Senegal. For all datasets, we standardized the transcriptions by lowercasing, removing punctuation, and converting digits to letters. The official train/test splits provided with Fleurs were retained, while for the other datasets we created random train/test partitions.
\\
\\
\myparagraph{Training.} To ensure broader usability, we converted the model from raw PyTorch weights into the HuggingFace format, which is the most widely adopted framework for fine-tuning open pretrained models. ASR Fine-tuning is performed using the official HuggingFace script for speech recognition using the CTC loss.
\\
\\
\myparagraph{Baselines.} We compare the model to Meta/HuBERT-Base \cite{hubert}, the model from which we started the continued pretraining. This helps to understand how much the continued pretraining improved the model. We also compare to Orange/HuBERT-Base \cite{afribert}, a HuBERT model trained from scratch on 65k hours of African languages speech, including Wolof. Both baseline models were fine-tuned on the same ASR dataset in Table \ref{tab:asr_data}

\myparagraph{Results.} Table \ref{tab:wer} shows that our model improves over Meta/HuBERT-Base and Orange/HuBERT-Base, despite being trained on far less speech data than the latter. We attribute this superior performance to continued pretraining (\textbf{Insight 1}) and to the Wolof-centric, high-quality data collection (\textbf{Insight 2}). Orange/HuBERT-Base underperforms compared to Meta/HuBERT-Base because the former is trained from scratch and probably with fewer epochs. This is also a reason to conduct continued pretraining: it requires fewer training steps to achieve good performance.

\begin{tcolorbox}[colback=violet!15!white,colframe=violet!100!white,title={\color{black} \textbf{Insight 1}: Always do continued pretraining when possible} ]
    Continued pretraining allows us to reduce training costs by leveraging the compute already spent in the base model pretraining, so only a few additional epochs are needed.
\end{tcolorbox}

\begin{tcolorbox}[colback=violet!15!white,colframe=violet!100!white,title={\color{black} \textbf{Insight 2}: Don't rely on multilingualism only, scale the monolingual dataset} ]
    Orange/HuBERT-base underperforms on Wolof despite 65k hours of multilingual training data with languages similar to Wolof. This suggests that cross-lingual transfer is insufficient, and Wolof-specific training on smaller, high-quality data can perform better.
\end{tcolorbox}
\begin{table}[ht]
    \centering
    \begin{tabular}{llc}
        \toprule
        \textbf{Model} & \textbf{Pretraining Hours} & \textbf{WER} ($\downarrow$) \\
        \midrule
        Orange/HuBERT-Base & 65 000 & \ 41.11 \\
        Meta/HuBERT-Base & 960 & 39.48 \\
        \rowcolor{violet!40} Ours & 960 $\rightarrow$ \textbf{860} & \textbf{35.65} \\
        \bottomrule
    \end{tabular}
    \caption{WER results of the continued pretrained HuBERT compared to the initial model (Meta/HuBERT-Base) and an African-centric model Orange/HuBERT-Base. 960 $\rightarrow$ \textbf{860} means the initial model is trained on 960 hours and continued pretrained on 860 hours of Wolof speech. In any case, the WER remains high because, contrary to English, \textbf{the orthography of Wolof is not standardized} since it's primarily an oral language.}
    \label{tab:wer}
\end{table}
\subsection{Discussion}
We have shown that Wolof-specific continued pretraining on an spontaneous, high-quality dataset can substantially improve speech processing tasks such as speech recognition. However, with only 90M parameters, its capacity remains limited for more complex tasks such as speech question answering or speech translation, which demand higher linguistic representations. In the next section, we integrate the Wolof HuBERT model with a Wolof language model to enhance speech recognition and extend its capabilities to speech translation. We study the integration of the Wolof HuBERT into an LLM. Since there is no open Wolof LLM, we first present the training of a text-based LLM for Wolof and then how we integrate the HuBERT encoder into the LLM.

\section{Speech Language Model}
\label{sec:slm}
\begin{figure*}[!ht]
    \begin{minipage}[l]{.48\linewidth}
        \centering
        \begin{tabular}{lcc}
            \toprule
            \textbf{CoT Order} & \textbf{WER} ($\downarrow$) & \textbf{CER} ($\downarrow$) \\
            \midrule
            \textsc{transcribe} & \textbf{29.09} & \textbf{15.26} \\
            \textsc{phonemize} $\rightarrow$ \textsc{transcribe} & 34.05 & 19.08 \\
            \textsc{translate} $\rightarrow$ \textsc{transcribe} & 29.48 & 15.95 \\
            \bottomrule
        \end{tabular}
        \captionof{table}{ASR Results of the Speech LLM. \textsc{x} $\rightarrow$ \textsc{y} means doing the CoT step \textsc{x} before performing the task \textsc{y}.}
        \label{tab:speechlm_asr}
    \end{minipage}
    \hfill
    \begin{minipage}[l]{.48\linewidth}
        \centering
        \begin{tabular}{lcc}
            \toprule
            \textbf{CoT Order} & \textbf{ChRF} ($\uparrow$) & \textbf{BS-F1} ($\uparrow$) \\
            \midrule
            \textsc{translate} & 33.08 & \textbf{79.78} \\
            \textsc{transcribe} $\rightarrow$ \textsc{translate} & \textbf{33.79} & 79.73 \\
            \textsc{reformulate} $\rightarrow$ \textsc{translate} & 33.59 & 77.54 \\
            \bottomrule
        \end{tabular}
        \captionof{table}{Speech LLM Translation Results. \textsc{x} $\rightarrow$ \textsc{y} means doing the CoT step \textsc{x} before performing the task \textsc{y}.}
        \label{tab:speechlm_translation}
    \end{minipage}
\end{figure*}

\subsection{Training Wolof LLM}
\myparagraph{Model.} We finetune the Wolof LLM from Qwen2.5 3B [cite]. While other LLMs can also be used, we chose this LLM not only for its good general performance but also for its open license, enabling us to redistribute the fine-tuned model freely. 
\\
\\
\myparagraph{Data.} We focus on tasks that enable cross-lingual transfer of the English knowledge encoded in the LLM. We train and optimize the model for bidirectional translation between English and Wolof. Additionally, we incorporate high-quality monolingual Wolof text, primarily sourced from Wolof-centric websites, to enhance natural language generation in this language. The resulting dataset contains about 2B tokens.

\subsection{Linking HuBERT to the LLM}
\myparagraph{Method.} There are many ways to integrate a speech encoder into an LLM. We adopt the \textit{late-fusion} \cite{llava, llamaspeech}, approach for its simplicity: speech features from the encoder and text embedding from the LLM's embedding layer are projected separately and then concatenated before being forwarded in the LLM. This method introduces minimal inference latency, as the speech features are projected only once through a lightweight Multi-Layer Perceptron (MLP). Figure~\ref{fig:speech_lm_approach} shows how we integrate HuBERT into the LLM. Features from all HuBERT layers are concatenated, projected by the speech aligner, then combined with the text embedding and fed to the LLM. Only the aligner is trained.
\\
\\
\myparagraph{Data with Multi-step Chain-of-Though.} We convert the ASR data in \ref{sec:asr_exps} into an instruction-following fine-tuning format, where the assistant is asked to transcribe or translate a given audio. We study the effect of Chain-of-Thought (CoT) on accuracy by adding to the training data multi-step CoT. More precisely, in addition to training the Speech LLM for direct transcription, we train it to \textsc{phonemize} then \textsc{transcribe} and to \textsc{translate} then \textsc{transcribe}. The intuition behind the phonemization step is to enable the LLM to construct a phonemic representation before transcription, while the translation step provides a semantic representation. We also add a multi-step CoT for the translation task. We train the Speech LLM to \textsc{transcribe} then \textsc{translate} or to \textsc{paraphrase} then \textsc{translate}. The intuition of the paraphrasing step before translation is to give the model more time to construct high-level speech understanding that may help for translation.
\\
\\
\myparagraph{Training.} We train the Speech LLM following the methodology and datasets described above. We use the CTC fine-tuned HuBERT from Section \ref{sec:asr_exps} as speech encoder, keeping it frozen during training along with the LLM. The speech–text aligner is a lightweight MLP with a single hidden layer and ReLU activation. No additional sequence downsampling is applied on the speech encoder output, since HuBERT already performs substantial downsampling of the audio. During training, the loss is only computed on the assistant completions. We use a small batch size of 2 and a learning rate of $1e-4$.

\subsection{Evaluation and Results}
We evaluate the Speech LLM on {\textbf{WER} for speech recognition. For speech-to-text translation, we use \textbf{ChRF}, a metric based on character n-gram matches. We complement this with a semantic-driven evaluation using BERTScore (\textbf{BS}), which uses a pretrained BERT model to compare candidate and reference translations via cosine similarity. We present the F1 score of BERTScore (\textbf{BS-F1})
\\
\\
\myparagraph{Result 1: The LLM improves speech recognition.} Table \ref{tab:speechlm_asr} shows that the Speech LLM achieves an 18.4\% improvement in WER compared to the HuBERT-only approach (see Table \ref{tab:wer}). Since the LLM was frozen during training, the results suggest that the improvement comes from the LLM providing additional Wolof linguistic knowledge.
\\
\\
\myparagraph{Result 2: The LLM enables speech translation.} Table \ref{tab:speechlm_translation} shows that the Speech LLM achieves good performance at speech-to-text translation. We found the model often provides paraphrase translations instead of literal ones, which explains why BERTScores (\textbf{BS-F1}) are better than \textbf{ChRF}
\\
\\
\myparagraph{Result 3: Multi-step CoT shows limited benefits.} Interestingly, multi-step CoT hasn't been beneficial in our case, probably due to the small model entering easily into a repetition loop, particularly when asked to phonemize first. We expect a large model to perform better with multi-step CoT.

\begin{tcolorbox}[colback=violet!15!white,colframe=violet!100!white,title={\color{black} \textbf{Insight 3}: Multi-step CoT doesn't always help} ]
While the LLM improves speech recognition and speech translation, doing extra CoT steps haven't been beneficial in our case, probably due to the small size of the model.
\end{tcolorbox}

\section{Conclusion}
We propose the first Speech LLM for Wolof and share insights relevant when working on underrepresented languages. We show the importance of scaling high-quality unsupervised monolingual datasets and of continued pretraining in achieving high-performing yet cost-efficient models. Our results show that integrating an LLM is beneficial for Wolof, improving speech recognition and enabling speech translation. For future work, we plan to scale the LLM size and investigate whether multi-step Chain-of-Thought can enhance speech processing tasks.

\bibliographystyle{IEEEbib}
\bibliography{refs}

\end{document}